%% file: journal.tex
\documentclass[journal]{IEEEtran}

\usepackage{footnote}
\makesavenoteenv{tabular}

\usepackage{amsthm}
\usepackage{float}
\usepackage{cite}								
\usepackage{amsmath}
\DeclareMathOperator{\pred}{Pred}
\DeclareMathOperator{\layerv}{Layer^v}
\DeclareMathOperator{\layerf}{Layer^f}
\DeclareMathOperator{\messageFN}{Message}
\DeclareMathOperator{\updateFN}{Update}
\DeclareMathOperator{\aggregateFN}{Aggregate}
\DeclareMathOperator{\attendFN}{Attend}

\usepackage[dvipsnames]{xcolor}
\usepackage{graphicx}
\usepackage{epstopdf}
\usepackage{float}
\usepackage{bm,upgreek}
\usepackage{amsfonts}
\usepackage{enumitem}
\usepackage{mathtools}
\usepackage{multirow}
\usepackage{booktabs}
\usepackage[subnum]{cases}
\usepackage{setspace}
\usepackage{color}
\usepackage{blkarray, bigstrut} 
\usepackage[caption=false,labelformat=simple]{subfig}

\usepackage{xcolor}
\usepackage{pgfplots}
\pgfplotsset{compat=1.5,
  tick label style={font=\small},
  label style={font=\small}}
\usepackage{pgfplotstable}
\usepgfplotslibrary{statistics}
\usepackage{tikz}

\usepackage{pgfplots}
\usepackage{pgfplotstable}
\pgfplotsset{compat=1.7}
\usepgfplotslibrary{groupplots}

\newcommand\scalemath[2]{\scalebox{#1}{\mbox{\ensuremath{\displaystyle #2}}}}   

\usepackage[toc,acronym]{glossaries}
\usepackage{mathtools,nccmath}
\usepackage[ruled]{algorithm}
\usepackage{algpseudocode,algorithmicx}

\algnewcommand{\LineComment}[1]{\State \(\#\) #1}

\allowdisplaybreaks

\ifCLASSINFOpdf
\else
\fi
\pgfplotsset{legend image with text/.style={legend image code/.code={%
\node[anchor=west, align=right] at (0.0cm,0cm) {#1};}},}

\makeatletter
\def\myline{\pgfutil@ifnextchar[{\my@line}{\my@line[]}}%
\def\my@line[#1](#2)(#3){%
\tikz[overlay] \draw[#1]  (#2)--(#3); 
}%
\algrenewcommand\algorithmicindent{1.0em}%

\makeatletter
\renewcommand{\ALG@beginalgorithmic}{\small}
\makeatother



\usepackage{etoolbox}
\BeforeBeginEnvironment{algorithmic}{\kern 0.3\baselineskip}
\AfterEndEnvironment{algorithmic}{\kern -0.1\baselineskip}

\makeatletter
\renewcommand{\ALG@beginalgorithmic}{\footnotesize}
\makeatother

\pgfplotsset{
    box plot/.style={
        /pgfplots/.cd,
        black,
        only marks,
        mark=-,
        mark size=\pgfkeysvalueof{/pgfplots/box plot width},
        /pgfplots/error bars/y dir=plus,
        /pgfplots/error bars/y explicit,
        /pgfplots/table/x index=\pgfkeysvalueof{/pgfplots/box plot x index},
    },
    box plot box/.style={
        /pgfplots/error bars/draw error bar/.code 2 args={%
            \draw  ##1 -- ++(\pgfkeysvalueof{/pgfplots/box plot width},0pt) |- ##2 -- ++(-\pgfkeysvalueof{/pgfplots/box plot width},0pt) |- ##1 -- cycle;
        },
        /pgfplots/table/.cd,
        y index=\pgfkeysvalueof{/pgfplots/box plot box top index},
        y error expr={
            \thisrowno{\pgfkeysvalueof{/pgfplots/box plot box bottom index}}
            - \thisrowno{\pgfkeysvalueof{/pgfplots/box plot box top index}}
        },
        /pgfplots/box plot
    },
    box plot top whisker/.style={
        /pgfplots/error bars/draw error bar/.code 2 args={%
            \pgfkeysgetvalue{/pgfplots/error bars/error mark}%
            {\pgfplotserrorbarsmark}%
            \pgfkeysgetvalue{/pgfplots/error bars/error mark options}%
            {\pgfplotserrorbarsmarkopts}%
            \path ##1 -- ##2;
        },
        /pgfplots/table/.cd,
        y index=\pgfkeysvalueof{/pgfplots/box plot whisker top index},
        y error expr={
            \thisrowno{\pgfkeysvalueof{/pgfplots/box plot box top index}}
            - \thisrowno{\pgfkeysvalueof{/pgfplots/box plot whisker top index}}
        },
        /pgfplots/box plot
    },
    box plot bottom whisker/.style={
        /pgfplots/error bars/draw error bar/.code 2 args={%
            \pgfkeysgetvalue{/pgfplots/error bars/error mark}%
            {\pgfplotserrorbarsmark}%
            \pgfkeysgetvalue{/pgfplots/error bars/error mark options}%
            {\pgfplotserrorbarsmarkopts}%
            \path ##1 -- ##2;
        },
        /pgfplots/table/.cd,
        y index=\pgfkeysvalueof{/pgfplots/box plot whisker bottom index},
        y error expr={
            \thisrowno{\pgfkeysvalueof{/pgfplots/box plot box bottom index}}
            - \thisrowno{\pgfkeysvalueof{/pgfplots/box plot whisker bottom index}}
        },
        /pgfplots/box plot
    },
    box plot median/.style={
        /pgfplots/box plot,
        /pgfplots/table/y index=\pgfkeysvalueof{/pgfplots/box plot median index},
        semithick,black
    },
    box plot width/.initial=1em,
    box plot x index/.initial=0,
    box plot median index/.initial=1,
    box plot box top index/.initial=2,
    box plot box bottom index/.initial=3,
    box plot whisker top index/.initial=4,
    box plot whisker bottom index/.initial=5,
}

\algnewcommand\algorithmicforeach{\textbf{for each}}
\algdef{S}[FOR]{ForEach}[1]{\algorithmicforeach\ #1\ \algorithmicdo}

\pgfplotsset{grid style={dotted,gray}}
\include{acronyms}

\begin{document}

\newlist{myitemize}{itemize}{3}
\setlist[myitemize,1]{label=\textbullet,leftmargin=6.5mm}

\title{Robust and Fast Data-Driven Power System State Estimator Using Graph Neural Networks}

\author{Ognjen~Kundacina,~\IEEEmembership{Student Member,~IEEE,}
        Mirsad~Cosovic,~\IEEEmembership{Member,~IEEE,}
        Dejan Vukobratovic,~\IEEEmembership{Senior Member,~IEEE}

\thanks{O. Kundacina is with The Institute for Artificial Intelligence Research and Development of Serbia (e-mail: ognjen.kundacina@ivi.ac.rs), M. Cosovic is with Faculty of Electrical Engineering, University of Sarajevo, Bosnia and Herzegovina (e-mail: mcosovic@etf.unsa.ba), D. Vukobratovic is with Faculty of Technical Sciences, University of Novi Sad, Serbia, (email: dejanv@uns.ac.rs).}}

\markboth{}%
{Shell \MakeLowercase{\textit{et al.}}: Bare Demo of IEEEtran.cls for IEEE Journals}

\maketitle

\begin{abstract}

The power system state estimation (SE) algorithm estimates the complex bus voltages based on the available set of measurements. Because phasor measurement units (PMUs) are becoming more widely employed in transmission power systems, a fast SE solver capable of exploiting PMUs' high sample rates is required. To accomplish this, we present a method for training a model based on graph neural networks (GNNs) to learn estimates from PMU voltage and current measurements, which, once it is trained, has a linear computational complexity with respect to the number of nodes in the power system. We propose an original GNN implementation over the power system's factor graph to simplify the incorporation of various types and numbers of measurements both on power system buses and branches. Furthermore, we augment the factor graph to improve the robustness of GNN predictions. Training and test examples were generated by randomly sampling sets of power system measurements and annotated with the exact solutions of linear SE with PMUs. The numerical results demonstrate that the GNN model provides an accurate approximation of the SE solutions. Additionally, errors caused by PMU malfunctions or the communication failures that make the SE problem unobservable have a local effect and do not deteriorate the results in the rest of the power system.

\end{abstract}


\IEEEpeerreviewmaketitle

\section{Introduction}

\textbf{Motivation:} The state estimation (SE), which estimates the set of power system state variables based on the available set of measurements, is an essential tool used for the power system's monitoring and operation \cite{monticelli2000SE}. With the increasing deployment of phasor measurement units (PMUs) in power systems, a fast state estimator is required to maximise the use of their high sampling rates. Solving PMU-based SE, represented as a linear weighted least-squares (WLS) problem, involves matrix factorisations. Neglecting the PMU measurement covariances represented in rectangular coordinates is a common practice \cite{exposito} that results in near $\mathcal{O}(n^2)$ computational complexity for sparse matrices, where $n$ denotes the number of power system's buses. Recent advancements in graph neural networks (GNNs) \cite{pmlr-v70-gilmer17a, GraphRepresentationLearningBook, velickovic2018graph} open up novel possibilities for developing power system algorithms with linear computational complexity and potential distributed implementation. One of the primary advantages of using GNNs in power systems instead of traditional deep learning methods is that the prediction model is not restricted to training and test examples of fixed power system topologies. GNNs take advantage of the graph structure of the input data, resulting in fewer learning parameters, lower memory requirements, and the incorporation of connectivity information into the learning process. Furthermore, the inference phase of the trained GNN model can be distributed and performed locally, since the prediction of the state variable of a node requires only $K$-hop neighbourhood measurements. 
Additional motivation for using GNNs as well as the other deep learning methods for data-driven SE comes from the power system parameters' uncertainty \cite{antona2018PSUncertainty}. However, note that GNNs provide an option of incorporating power system parameters via edge input features \cite{gong2019EdgeFeatures}.

\textbf{Literature Review:} 
Several studies suggest using a deep learning model to learn the solutions to computationally demanding algorithms for power system analysis. In \cite{zhang2019}, combination of recurrent and feed-forward neural networks is used to solve the power system SE problem, given the measurement data and the history of network voltages. An example of training a feed-forward neural network to initialise the network voltages for the Gauss-Newton distribution system SE solver is given in \cite{zamzam2019}. GNNs are beginning to be used for solving similar problems, like in \cite{donon2019graphneuralsolver} and \cite{bolz2019PFapproximator}, where power flows in the system are predicted based on the power injection data labelled by the traditional power flow solver. Similarly, \cite{wang2020ProbabPF} proposes using a trained GNN as an alternative to computationally expensive probabilistic power flow, which calculates probability density functions of the unknown variables. Different approach is suggested in \cite{donon2020graphneuralsolverJOURNAL}, where a GNN is trained to perform the power flow calculation by minimizing the violation of Kirchhoff’s law at each bus, avoiding the requirement for having data labeled by a conventional power flow solver. In \cite{pagnier2021physicsinformed}, the authors propose a combined model and data-based approach in which GNNs are used for the power system parameter and SE. The model predicts power injections/consumptions in the nodes where voltages and phases are measured, whereas it does not include the branch measurements, as well as node measurement types other than the voltage in the calculation. In \cite{Yang2022RobustPSSEDataDrivenPriors}, GNN was trained by propagating simulated or measured voltages through the graph to learn the voltage labels from the historical dataset and used as a part of the regularization term in the SE loss function. However, the proposed GNN also does not take into account measurement types other than node voltage measurements, though they are handled in other parts of the algorithm. An additional feed-forward neural network was trained to learn the solutions that minimise the SE loss function, resulting in accelerated SE solution with $\mathcal{O}(n^2)$ computational complexity during inference. In \cite{TGCN_SE_2021}, state variables are predicted based on a time-series of node voltage measurements, and the SE problem is solved by employing GNNs enriched with gated recurrent units.

\textbf{Contributions:} This paper proposes training a GNN for a regression problem using the inputs and the solutions of linear SE with PMUs to provide fast and accurate predictions during the evaluation phase. The following are the paper's main contributions:
\begin{itemize}[leftmargin=*]
\item Inspired by \cite{Satorras2021NeuralEB}, we present the first attempt to utilise GNNs over the factor graphs \cite{Kschischang2001FactorGraphs} instead of the bus-branch power system model for the SE problem. This enables trivial integration and exclusion of any type and number of measurements on the power system's buses and branches, by adding or removing the corresponding nodes in the factor graph. Moreover, the factor graph is augmented by adding direct connections between variable nodes to improve information propagation during neighborhood aggregation, particularly in unobservable scenarios when the loss of the measurement data occurs.
\item We present a graph attention network (GAT) \cite{velickovic2018graph} based model to solve the SE problem, with the architecture specialised for the proposed heterogeneous augmented factor graph. GNN layers that aggregate into factor and variable nodes each have their own set of trainable parameters. Furthermore, distinct sets of parameters are used for variable-to-variable and factor-to-variable message functions in GNN layers that aggregate into variable nodes.
\item Due to the sparsity of the power system graph, and the fact that the node degree does not increase with the total number of nodes, the proposed approach has $\mathcal{O}(n)$ computational complexity, making it suitable for application in large-scale power systems.
\item We investigate the proposed model's local-processing nature and demonstrate that significant result degradation caused by PMU or communication failures affects only the local neighborhood of the node where the failure occurred.
\item We examine the proposed model's robustness to outliers and improve it solely by augmenting the training set with examples that contain outliers.
\item The inference of the trained GNN is easy to distribute and parallelize. Even in the case of centralised SE implementation, the processing can be done using distributed computation resources, such as graphical-processing units.
\end{itemize}


We note that this paper significantly extends the conference version \cite{kundacina2022state}. The GNN model architecture is additionally improved and described in greater detail, and the algorithm's performance and scalability are enhanced by using variable node index binary encoding. GNN inputs now additionally include measurement covariances represented in rectangular coordinates, and discussions on robustness to outliers and computational complexity are added. We extended the numerical results section with experiments that include additional cases of measurement variances and redundancies and tested the algorithm's scalability on larger power systems with sizes ranging from 30 to 2000 buses. In addition, we compared all of the results with the approximate linear WLS solution of SE with PMUs, in which covariances of measurement represented in rectangular coordinates are neglected.

The rest of the paper is organised as follows. Section II provides a formulation of the linear SE with PMUs problem. Section III gives the theoretical foundations of GNNs and defines the augmented factor graph and describes the GNN architecture. Section IV shows numerical results from various test scenarios, while Section V draws the main conclusions of the work.

\section{Linear State Estimation with PMUs}
\label{sec:lse}

The SE algorithm estimates the values of the state variables $\mathbf{x}$ based on the knowledge of the network topology and parameters, and measured values obtained from the measurement devices spread across the power system. 

The power system network topology is described by the bus/branch model and can be represented using a graph $\mathcal{G} =(\mathcal{H},\mathcal{E})$, where the set of nodes $\mathcal{H} = \{1,\dots,n \}$ represents the set of buses, while the set of edges $\mathcal{E} \subseteq \mathcal{H} \times \mathcal{H}$ represents the set of branches of the power network. The branches of the network are defined using the two-port $\pi$-model. More precisely, the branch $(i,j) \in \mathcal{E}$ between buses $\{i,j\} \in \mathcal{H}$ can be modelled using complex expressions:
\begin{equation}
  \begin{bmatrix}
    {I}_{ij} \\ {I}_{ji}
  \end{bmatrix} =
  \begin{bmatrix}
    \cfrac{1}{\tau_{ij}^2}(y_{ij} + y_{\text{s}ij}) & -\alpha_{ij}^*{y}_{ij}\\
    -\alpha_{ij}{y}_{ij} & {y}_{ij} + y_{\text{s}ij}
  \end{bmatrix}  
  \begin{bmatrix}
    {V}_{i} \\ {V}_{j}
  \end{bmatrix},
  \label{unified}
\end{equation} 
where the parameter $y_{ij} = g_{ij} + \text{j}b_{ij}$ represents the branch series admittance, half of the total branch shunt admittance (i.e., charging admittance) is given as $y_{\text{s}ij} = \text{j}b_{si}$. Further, the transformer complex ratio is defined as $\alpha_{ij} = (1/\tau_{ij})\text{e}^{-\text{j}\phi_{ij}}$, where $\tau_{ij}$ is the transformer tap ratio magnitude, while $\phi_{ij}$ is the transformer phase shift angle. It is important to remember that the transformer is always located at the bus $i$ of the branch described by \eqref{unified}. Using the branch model defined by \eqref{unified}, if $\tau_{ij} = 1$ and $\phi_{ij} = 0$ the system of equations describes a line. In-phase transformers are defined if $\phi_{ij} = 0$ and $y_{\text{s}ij} = 0$, while phase-shifting transformers are obtained if $y_{\text{s}ij} = 0$. The complex expressions ${I}_{ij}$ and ${I}_{ji}$ define branch currents from the bus $i$ to the bus $j$, and from the bus $j$ to the bus $i$, respectively. The complex bus voltages at buses $\{i,j\}$ are given as $V_i$ and $V_j$, respectively.  

PMUs measure complex bus voltages and complex branch currents. More precisely, phasor measurement provided by PMU is formed by a magnitude, equal to the root mean square value of the signal, and phase angle \cite[Sec.~5.6]{phadke}. The PMU placed at the bus measures bus voltage phasor and current phasors along all branches incident to the bus \cite{exposito}. Thus, the PMU outputs phasor measurements in polar coordinates. In addition, PMU outputs can be observed in the rectangular coordinates with real and imaginary parts of the bus voltage and branch current phasors. In that case, the vector of state variables $\mathbf{x}$ can be given in rectangular coordinates $\mathbf x \equiv[\mathbf{V}_\mathrm{re},\mathbf{V}_\mathrm{im}]^{\mathrm{T}}$, where we can observe real and imaginary components of bus voltages as state variables:   
\begin{equation}
   	\begin{aligned}
        \mathbf{V}_\mathrm{re}&=\big[\Re({V}_1),\dots,\Re({V}_n)\big]\\
	    \mathbf{V}_\mathrm{im}&=\big[\Im({V}_1),\dots,\Im({V}_n)\big].     
   	\end{aligned}
   	\label{rect_coord}
\end{equation} 

Using rectangular coordinates, we obtain the linear system of equations defined by voltage and current measurements obtained from PMUs. The measurement functions corresponding to the bus voltage phasor measurement on the bus $i \in \mathcal{H}$ are simply equal to: 
\begin{equation}
    \begin{aligned}
        f_{\Re\{V_i\}}(\cdot) = \Re\{V_i\}\\
        f_{\Im\{V_i\}}(\cdot) = \Im\{V_i\}.
    \end{aligned}  
    \label{measFunction1}
\end{equation}
According to the unified branch model \eqref{unified}, functions corresponding to the branch current phasor measurement vary depending on where the PMU is located. If PMU is placed at the bus $i$, functions are given as:
\begin{equation}
\scalemath{0.83}{
    \begin{aligned}
        f_{\Re\{I_{ij}\}}(\cdot) &= q \Re\{V_{i}\} - w \Im\{V_{i}\} - (r-t) \Re\{V_{j}\} + (u+p) \Im\{V_{j}\} \\
        f_{\Im\{I_{ij}\}}(\cdot) &= w \Re\{V_{i}\} + q \Im\{V_{i}\} - (u+p) \Re\{V_{j}\} - (r-t) \Im\{V_{j}\},
    \end{aligned}}
    \label{measFunction2}
\end{equation}
where $q =$ $g_{ij}/\tau_{ij}^2$, $w =$ $(b_{ij} + b_{si})/\tau_{ij}^2$, $r =$ $(g_{ij}/\tau_{ij})$ $\cos\phi_{ij}$, $t =$ $(b_{ij}/\tau_{ij})$ $\sin\phi_{ij}$, $u =$ $(b_{ij}/\tau_{ij})$ $\cos\phi_{ij}$, $p =$ $(g_{ij}/\tau_{ij})$ $\sin\phi_{ij}$. In the case where PMU is installed at the bus $j$, measurement functions are: 
\begin{equation}
\scalemath{0.83}{
    \begin{aligned}
        f_{\Re\{I_{ji}\}}(\cdot) &= z \Re\{V_{j}\} - e \Im\{V_{j}\} - (r+t) \Re\{V_{i}\} + (u-p) \Im\{V_{i}\} \\
        f_{\Im\{I_{ji}\}}(\cdot) &= e \Re\{V_{j}\} + z \Im\{V_{j}\} - (u-p) \Re\{V_{i}\} - (r+t) \Im\{V_{i}\},
    \end{aligned}}
    \label{measFunction3}
\end{equation}
where $z = g_{ij}$ and $e =$ $b_{ij} + b_{si}$. The presented model represents the system of linear equations, where the solution can be found by solving the linear weighted least-squares problem: 
\begin{equation}
    \left(\mathbf H^{T} \mathbf \Sigma^{-1} \mathbf H \right) \mathbf x =
		\mathbf H^{T} \mathbf \Sigma^{-1} \mathbf z,    
	\label{SE_system_of_lin_eq}
\end{equation}
where the Jacobian matrix $\mathbf {H} \in \mathbb {R}^{k \times 2n}$ is defined according to measurement functions \eqref{measFunction1}-\eqref{measFunction3}, $k$ is the total number of linear equations, the observation error covariance matrix is given as $\mathbf {\Sigma} \in \mathbb {R}^{k \times k}$, and the vector $\mathbf z \in \mathbb {R}^{k}$ contains measurement values given in rectangular coordinate system. 

The main disadvantage of this approach is that  measurement errors correspond to polar coordinates (i.e., magnitude and angle errors), whereas the covariance matrix must be transformed from polar to rectangular coordinates \cite{zhou2016phasorsMeasSE}. As a result, measurement errors are correlated and covariance matrix $\mathbf {\Sigma}$ does not have a diagonal form. Despite that, because of the lower computational effort, the non-diagonal elements of the covariance matrix $\mathbf {\Sigma}$ are usually neglected, which has an effect on the accuracy of the state estimation \cite{exposito}. Using the classical theory of propagation of uncertainty, the variance in the rectangular coordinate system can be obtained using variances in the polar coordinate system. For example, let us observe the voltage phasor measurement at the bus $i$, where PMU outputs the voltage magnitude measurement value $z_{|V_i|}$ with corresponding variance $v_{|V_i|}$, and voltage phase angle measurement $z_{\theta_i}$ with variance $v_{\theta_i}$. Then, variances in the rectangular coordinate system can be obtained as:
\begin{equation}
    \begin{aligned}
        v_{\Re\{V_i\}} &= v_{|V_i|} (\cos z_{\theta_i})^2 + v_{\theta_i} (z_{|V_i|} \sin z_{\theta_i})^2\\
        v_{\Im\{V_i\}} &= v_{|V_i|} (\sin z_{\theta_i})^2 + v_{\theta_i} (z_{|V_i|} \cos z_{\theta_i})^2.
    \end{aligned}
\end{equation}
Analogously, we can easily compute variances related to current measurements $v_{\Re\{I_{ij}\}}$, $v_{\Im\{I_{ij}\}}$ or $v_{\Re\{I_{ji}\}}$, $v_{\Im\{I_{ji}\}}$. 
We will refer to the solution of \eqref{SE_system_of_lin_eq} in which measurement error covariances are neglected to avoid the computationally demanding inversion of the non-diagonal matrix $\mathbf {\Sigma}$ as an \textit{approximative WLS SE solution}. 

In this work, we will research if the GNN model trained with measurement values, variances, and covariances labeled with the exact solutions of \eqref{SE_system_of_lin_eq} is more accurate that the approximative WLS SE, which neglects the covariances. Inference performed using the trained GNN model scales with linear computational complexity regarding the number of power system buses, making it significantly faster than both the approximate and the exact solver of \eqref{SE_system_of_lin_eq}.

\section{Graph Neural Network-based State Estimation}
In this section, we present the GNN theory foundations, augmentation techniques for the power system's factor graph, the details of the proposed GNN architecture, and analyse the computational complexity and the distributed implementation of the GNN model's inference.

\subsection{Basics of Graph Neural Networks} 

\begin{figure}[!t]
\centering

\begin{tikzpicture} [scale=0.71, transform shape]

\tikzset{
    box/.style={draw, fill=Goldenrod, minimum width=1.5cm, minimum height=0.8cm}}
    
\begin{scope}[local bounding box=graph]

\node [box]  (message1) at (0, 1.5) {Message};
\node [box]  (message2) at (0, 0) {Message};
\node [box]  (gat) at (3, 0) {$\aggregateFN$};
\node [box]  (update) at (5.5, 0) {$\updateFN$};


\draw[-stealth] (-1.8, 0) -- (message2.west) node[at start,left]{${h_{2}}^{k-1}$};

\draw[-stealth] (-1.8, 1.5) -- (message1.west) node[at start,left]{${h_{1}}^{k-1}$};

\draw[-stealth] (message1.east) -- (gat.west);

\draw[-stealth] (message1.east) -- (gat.west);
	
\draw[-stealth] (message2.east) -- (gat.west);
	
\draw[-stealth] (2, -1) -- (gat.west)
	node[at start,left]{$...$} node[very near end,left]{$...$};
	
\draw[-stealth] (gat.east) -- (update.west);

\draw[-stealth] (update.east) -- (7.5, 0) node[at end,right]{${h_{j}}^{k}$};

\end{scope}

\end{tikzpicture}

\caption{The detailed structure of one GNN layer, where yellow rectangles represent functions with trainable parameters.}
    \label{GNNlayerDetails}
\end{figure}
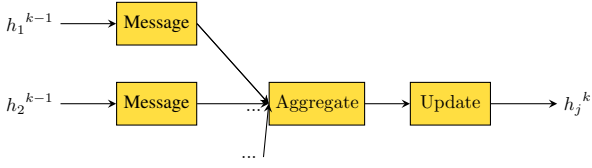

The spatial GNNs used in this study learn over graph-structured data using a recursive neighbourhood aggregation scheme, also known as the message passing procedure \cite{pmlr-v70-gilmer17a}. This results in an $s$-dimensional vector embedding $\mathbf h \in \mathbb {R}^{s}$ of each node, which captures the information about the node's position in the graph, as well as it's own and the input features of the neighboring nodes. The GNN layer, which implements one iteration of the recursive neighbourhood aggregation consists of several functions, that can be represented using a trainable set of parameters, usually in form of the feed-forward neural networks. One of the functions is the message function $\messageFN(\cdot|\theta^{\messageFN}): \mathbb {R}^{2s} \mapsto \mathbb {R}^{u}$, which outputs the message $\mathbf m_{i,j} \in \mathbb {R}^{u}$ between two node embeddings, $\mathbf h_i$ and $\mathbf h_j$. The second one is the aggregation function $\aggregateFN(\cdot|\theta^{\aggregateFN}): \mathbb {R}^{\textrm{deg}(j) \cdot u} \mapsto \mathbb {R}^{u}$, which defines in which way are incoming neighboring messages combined, which outputs the aggregated messages denoted as $\mathbf {m_j} \in \mathbb {R}^{u}$ for node $j$. The output of one iteration of neighbourhood aggregation process is the updated node embedding obtained by applying the update function $\updateFN(\cdot|\theta^{\updateFN}): \mathbb {R}^{u} \mapsto \mathbb {R}^{s}$ on the aggregated messages. The recursive neighbourhood aggregation process is repeated a predefined number of iterations $K$, also known as the number of GNN layers, where the initial node embedding values are equal to the $l$-dimensional node input features, linearly transformed to the initial node embedding $\mathbf {h_j}^0 \in \mathbb {R}^{s}$. The iteration that the node embeddings and calculated messages correspond to is indicated by the superscript. One iteration of the neighbourhood aggregation process for the $k^{th}$ GNN layer is depicted in Fig.~\ref{GNNlayerDetails} and also described by equations \eqref{gnn_equations}:
\begin{equation}
    \begin{gathered}
        \mathbf {m_{i,j}}^{k-1} = \messageFN( \mathbf {h_i}^{k-1}, \mathbf {h_j}^{k-1} | \theta^{\messageFN})\\
        \mathbf {m_j}^{k-1} = \aggregateFN(\{{\mathbf m_{i,j}}^{k-1} | i \in \mathcal{N}_j\} | \theta^{\aggregateFN})\\
        \mathbf {h_j}^k = \updateFN(\mathbf {m_j}^{k-1} | \theta^{\updateFN})\\
        k \in \{1,\dots,K\},
    \end{gathered}
    \label{gnn_equations}
\end{equation}
where $\mathcal{N}_j$ denotes the $1$-hop neighbourhood of the node $j$. The output of the whole process are final node embeddings $\mathbf {h_j}^{K}$ which can be used for the classification or regression over the nodes, edges, or the whole graph, or can be used directly for the unsupervised node or edge analysis of the graph. In the case of supervised learning over the nodes, the final embeddings are passed through the additional nonlinear function, creating the outputs that represent the predictions of the GNN model for the set of inputs fed into the nodes and their neighbours. The GNN model is trained by backpropagation of the loss function between the labels and the predictions over the whole computational graph. We refer the reader to \cite{GraphRepresentationLearningBook} for a more detailed introduction to graph representation learning and GNNs.

\subsection {Graph attention networks}
An important decision to be made while creating the GNN model is to select the GNN's aggregation function. Aggregation functions that are commonly used include sum, mean, min and max pooling, and graph convolution \cite{KipfW17_GCN}. One common drawback of these approaches is that incoming messages from all of the node's neighbours are weighted equally, or using weights calculated using the structural properties of the graph (e.g., node degrees) prior to training. Graph attention networks (GATs) \cite{velickovic2018graph} propose using the attention-based aggregation, in which the weights that correspond to the importance of each neighbour's message are learned from the data.
 
The attention mechanism introduces an additional set of trainable parameters in the aggregation function, usually implemented as a feed-forward neural network $\attendFN(\cdot|\theta^{\attendFN}): \mathbb {R}^{2s} \mapsto \mathbb {R}$ applied over the concatenated initial embeddings of the node $j$ and each of its neighbours:
 
\begin{equation}
    e_{i,j} = \attendFN(\mathbf {h_i}^{0}, \mathbf {h_j}^{0} | \theta^{\attendFN}).
    \label{unnormalized_attention}
\end{equation}

We obtain the final attention weights $a_{i,j} \in \mathbb {R}$ by normalising the output $e_{i,j} \in \mathbb {R}$ using the softmax function:
 \begin{equation}
    a_{i,j} = \frac{exp({e_{i,j}})}{\sum_{i' \in \mathcal{N}_j} exp({e_{i',j}})}.
    \label{normalized_attention}
\end{equation}

\subsection{Power System factor graph augmentation and the proposed GNN architecture} 
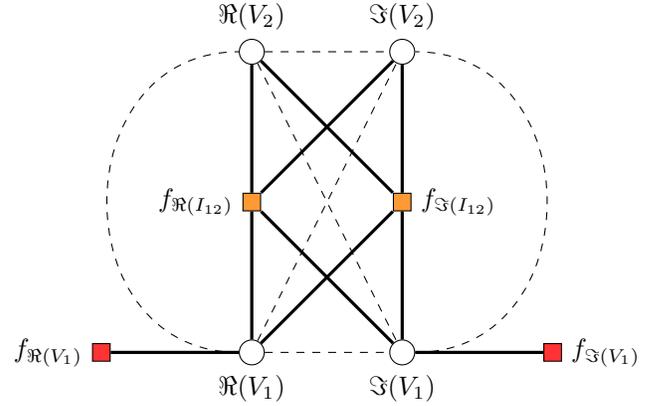
\begin{figure}[htbp]
    \centering
    \begin{tikzpicture} [scale=1.0, transform shape]
        \tikzset{
            varNode/.style={circle,minimum size=2mm,fill=white,draw=black},
            factorVoltage/.style={draw=black,fill=red!80, minimum size=2mm},
            factorCurrent/.style={draw=black,fill=orange!80, minimum size=2mm},
            edge/.style={very thick,black},
            edge2/.style={dashed,black}}
        \begin{scope}[local bounding box=graph]
            \node[factorVoltage, label=left:$f_{\Re({V}_1)}$] (f1) at (-3, 1 * 2) {};
            \node[factorVoltage, label=right:$f_{\Im({V}_1)}$] (f4) at (3, 1 * 2) {};
            \node[varNode, label=below:$\Re({V}_1)$] (v1) at (-1, 1 * 2) {};
            \node[varNode, label=below:$\Im({V}_1)$] (v3) at (1, 1 * 2) {};
            \node[factorCurrent, label=left:$f_{\Re({I}_{12})}$] (f2) at (-1, 2 * 2) {};
            \node[factorCurrent, label=right:$f_{\Im({I}_{12})}$] (f5) at (1, 2 * 2) {};
            \node[varNode, label=above:$\Re({V}_2)$] (v2) at (-1, 3 * 2) {};
            \node[varNode, label=above:$\Im({V}_2)$] (v4) at (1, 3 * 2) {};
            
            \draw[edge] (f1) -- (v1);
            \draw[edge] (f4) -- (v3);
            \draw[edge] (f2) -- (v1);
            \draw[edge] (f5) -- (v3);
            \draw[edge] (f5) -- (v1);
            \draw[edge] (f2) -- (v3);
            \draw[edge] (f2) -- (v2);
            \draw[edge] (f2) -- (v4);
            \draw[edge] (f5) -- (v2);
            \draw[edge] (f5) -- (v4);
            \draw[edge2] (v1) to [out=180,in=180,looseness=1.5] (v2);
            \draw[edge2] (v1) -- (v3);
            \draw[edge2] (v1) -- (v4);
            \draw[edge2] (v2) -- (v3);
            \draw[edge2] (v2) -- (v4);
            \draw[edge2] (v3) to [out=0,in=0,looseness=1.5] (v4);
            
        \end{scope}
    \end{tikzpicture}
    \caption{Example of the factor graph (full-line edges) and the augmented factor graph for a two-bus power system. Variable nodes are depicted as circles, and factor nodes are as squares, colored differently to distinguish between measurement types.}
    \label{toyFactorGraph}
    
\end{figure}

Inspired by the recent work \cite{cosovic2019bpse} in which the power system SE is modelled as a probabilistic graphical model, the initial version of the graph over which GNN operates has a power system's factor graph topology, which is a bipartite graph that consists of the factor and variable nodes and the edges between them. Variable nodes are used to generate an s-dimensional node embedding for all of the state variables defined in \eqref{rect_coord}, i.e., real and imaginary parts of the bus voltages, $\Re({V}_i)$ and $\Im({V}_i)$. Factor nodes, two per each measurement phasor, serve as inputs for the measurement values, variances, and covariances, also given in rectangular coordinates, and whose embedded values are sent to variable nodes via GNN message passing. Note that by adding the measurement covariances to the GNN inputs, we provide the predictive model with the information which is neglected in the approximative WLS SE solution. Feature augmentation using binary index encoding is performed for variable nodes only, to help the GNN model represent the neighbourhood structure of a node better since variable nodes have no additional input features. Using binary encoding, as opposed to the one-hot encoding suggested in \cite{kundacina2022state}, significantly reduces the number of input neurons of the GNN layers, reducing the overall number of trainable parameters as well as training and inference time. Nodes connected by full lines in Fig.~\ref{toyFactorGraph} represent a simple example of the factor graph for a two-bus power system, with a PMU on the first bus, containing one voltage and one current phasor measurement. Compared to the approaches like \cite{pagnier2021physicsinformed}, in which GNN nodes correspond to state variables only, we find factor-graph-like GNN topology convenient for incorporating measurements in the GNN, because factor nodes can be added or removed from any place in the graph, using which one can simulate inclusion of various types and quantities of measurements both on power system buses and branches.

Augmenting the factor graph topology by connecting the variable nodes in the $2$-hop neighbourhood significantly improves the model's prediction quality in unobservable scenarios \cite{kundacina2022state}. The idea behind this proposal is that the graph should stay connected even in a case of simulating measurement loss by removing the factor nodes, enabling the messages to still be propagated in the whole $K$-hop neighbourhood of the variable node. In other words, a factor node corresponding to a branch current measurement can be removed, while still preserving the physical connection that exists between the power system buses. This requires adding an additional set of trainable parameters for the variable-to-variable message function. We will still use the terms factor and variable nodes, although the augmented factor graph displayed in Fig.~\ref{toyFactorGraph} with both full and dashed lines, is not a factor graph since it is no longer bipartite.

\begin{figure}[!t]
\centering
\subfloat[]{
\begin{tikzpicture} [scale=0.7, transform shape]

\tikzset{
    varNode/.style={circle,minimum size=6mm,fill=white,draw=black},
    factorVoltage/.style={draw=black,fill=red!80, minimum size=6mm},
    factorCurrent/.style={draw=black,fill=orange!80, minimum size=6mm},
    box/.style={draw, fill=Goldenrod, minimum width=1.5cm, minimum height=0.9cm}}

\begin{scope}[local bounding box=graph]

\node [box]  (layerF0) at (-3, 1.5) {$\layerf$};
\node [box]  (layerV0) at (-3, 0) {$\layerv$};

\node[factorVoltage, label=above:${h_{f}}^{K-1}$] (facReV1) at (0, 1.5) {};
\node[varNode, label=below:${h_{v2}}^{K-1}$] (varImV1) at (0, 0) {};
\node[varNode, label=above:${h_{v}}^{K}$] (varReV1) at (6, 0) {};

\node [box]  (layerV1) at (3, 0) {$\layerv$};
\node [box]  (pred) at (6, -3) {$\pred$};
\node [box, fill=SpringGreen]  (loss) at (3, -3) {Loss};

\draw[-stealth] (-4.25, 1.75) -- (layerF0.west);
\draw[-stealth] (-4.25, 1.5) -- (layerF0.west) node[at start,left]{$...$};
\draw[-stealth] (-4.25, 1.25) -- (layerF0.west);

\draw[-stealth] (-4.25, 0.25) -- (layerV0.west);
\draw[-stealth] (-4.25, 0) -- (layerV0.west) node[at start,left]{$...$};
\draw[-stealth] (-4.25, -0.25) -- (layerV0.west);

\draw[-stealth] (layerF0.east) -- (facReV1.west);
	
\draw[-stealth] (layerV0.east) -- (varImV1.west);
	
\draw[-stealth] (varImV1.east) -- (layerV1.west);
	
\draw[-stealth] (0.25, -2.5) -- (layerV1.west)
	node[at start,left]{$...$} node[very near end,left]{$...$};
	
\draw[-stealth] (facReV1.east) -- (layerV1.west);
	
\draw[-stealth] (layerV1.east) -- (varReV1.west);
	
\draw[-stealth] (varReV1.south) -- (pred.north);
	
\draw[-stealth] (pred.west) -- (loss.east)
	node[midway,above]{$output$};
	
\draw[-stealth] (3, -2.1) -- (loss.north)
	node[at start,above]{$label$};

\end{scope}

\end{tikzpicture}
}
\hfil
\subfloat[]{

\begin{tikzpicture} [scale=0.68, transform shape]

\tikzset{
    box/.style={draw, fill=Goldenrod, minimum width=1.5cm, minimum height=0.8cm}}
    
\begin{scope}[local bounding box=graph]

\node [box]  (message1) at (0, 1.5) {Message\textsuperscript{f$\rightarrow$v}};
\node [box]  (message2) at (0, 0) {Message\textsuperscript{v$\rightarrow$v}};
\node [box]  (gat) at (3, 0) {GAT};
\node [box]  (update) at (5.5, 0) {$\updateFN$};


\draw[-stealth] (-1.8, 0) -- (message2.west) node[at start,left]{${h_{v2}}^{K-1}$};

\draw[-stealth] (-1.8, 1.5) -- (message1.west) node[at start,left]{${h_{f}}^{K-1}$};

\draw[-stealth] (message1.east) -- (gat.west);

\draw[-stealth] (message1.east) -- (gat.west);
	
\draw[-stealth] (message2.east) -- (gat.west);
	
\draw[-stealth] (2, -1) -- (gat.west)
	node[at start,left]{$...$} node[very near end,left]{$...$};
	
\draw[-stealth] (gat.east) -- (update.west);

\draw[-stealth] (update.east) -- (7.5, 0) node[at end,right]{${h_{v}}^{K}$};


\end{scope}

\label{layerDetails}

\end{tikzpicture}
}
\caption{Subfigure (a) illustrates a high-level computational graph that begins with the loss function for the output of a variable node $v$. Subfigure (b) depicts the detailed structure of one GNN $\layerv$, while yellow rectangles represent functions with trainable parameters.}
    \label{computationalGraph}
\end{figure}
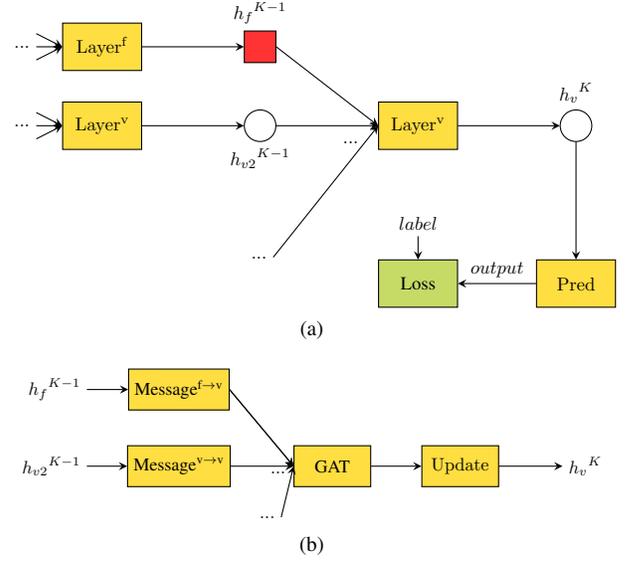

Since our GNN operates on a heterogeneous graph, in this work we employ two different types of GNN layers, $\layerf(\cdot|\theta^{\layerf}): \mathbb {R}^{\textrm{deg}(f)  \cdot s} \mapsto \mathbb {R}^{s}$ for aggregation in factor nodes $f$, and $\layerv(\cdot|\theta^{\layerv}): \mathbb {R}^{\textrm{deg}(v) \cdot s} \mapsto \mathbb {R}^{s}$ for variable nodes $v$, so that their message, aggregation, and update functions are learned using a separate set of trainable parameters, denoted all together as  $\theta^{\layerf}$ and $\theta^{\layerv}$. Additionally, we use a different set of trainable parameters for variable-to-variable and factor-to-variable node messages, $\textrm{Message}\textsuperscript{f$\rightarrow$v}(\cdot|\theta^{\textrm{Message}\textsuperscript{f$\rightarrow$v}}): \mathbb {R}^{2s} \mapsto \mathbb {R}^{u}$ and $\textrm{Message}\textsuperscript{v$\rightarrow$v}(\cdot|\theta^{\textrm{Message}\textsuperscript{v$\rightarrow$v}}): \mathbb {R}^{2s} \mapsto \mathbb {R}^{u}$ , in the $\layerv(\cdot|\theta^{\layerv})$ layer. In both GNN layers, we used two-layer feed-forward neural networks as message functions, single layer neural networks as update functions and the attention mechanism in the aggregation function. Furthermore, we apply a two-layer neural network $\pred(\cdot|\theta^{\pred}): \mathbb {R}^{s} \mapsto \mathbb {R}$ on top of the final node embeddings $\mathbf h^K$ of variable nodes only, to create the state variable predictions $\mathbf{x^{pred}}$. For the factor and variable nodes with indices $f$ and $v$, neighbourhood aggregation, and state variable prediction can be described as: 
\begin{equation}
    \begin{gathered}
        \mathbf {h_v}^k = \layerv(\{\mathbf {h_i}^{k-1} | i \in \mathcal{N}_v\} | \theta^{\layerv})\\
        \mathbf {h_f}^k = \layerf(\{\mathbf {h_i}^{k-1} | i \in \mathcal{N}_f\} | \theta^{\layerf})\\
        {x_v}^{pred} = \pred(\mathbf {h_v}^K|\theta^{\pred})\\
        k \in \{1,\dots,K\},
    \end{gathered}
    \label{embeddingds_and_predictions}
\end{equation}
where $\mathcal{N}_v$ and $\mathcal{N}_f$ denote the $1$-hop neighbourhoods of the nodes $v$ and $f$. All of the GNN trainable parameters $\theta$ are updated by applying gradient descent (i.e., backpropagation) to a loss function calculated over the whole mini-batch of graphs, as a mean squared difference between the state variable predictions and labels $\mathbf{x^{label}}$:
\begin{equation} \label{loss_function}
    \begin{gathered}
        L(\theta) = \frac{1}{2nB} \sum_{i=1}^{2nB}({{x_i}^{pred}} - {{x_i}^{label}})^2 \\
        \theta = \{\theta^{\layerv} \mathop{\cup} \theta^{\layerf} \mathop{\cup} \theta^{\pred}\} \\
        \theta^{\layerv} = \{\theta^{\textrm{Message}\textsuperscript{f$\rightarrow$v}} \mathop{\cup} \theta^{\textrm{Message}\textsuperscript{v$\rightarrow$v}} \mathop{\cup} \theta^{\textrm{Aggregate}\textsuperscript{v}} \mathop{\cup} \theta^{\textrm{Update}\textsuperscript{v}}\}\\
        \theta^{\layerf} = \{\theta^{\textrm{Message}\textsuperscript{v$\rightarrow$f}} \mathop{\cup} \theta^{\textrm{Aggregate}\textsuperscript{f}} \mathop{\cup} \theta^{\textrm{Update}\textsuperscript{f}}\},
    \end{gathered}
\end{equation}
where $2n$ is the total number of variable nodes in a graph, and $B$ is the number of graphs in the mini-batch. Fig.~\ref{computationalGraph} shows the high-level computational graph for the output of a variable node from the augmented factor graph given in Fig.~\ref{toyFactorGraph}. For simplicity, we present only one unrolling of the neighbourhood aggregation, as well as the details of $\theta^{\layerv}$ only.

\subsection{Computational complexity and distributed inference} 
Because the node degree in the graph of the power system does not increase with the total number of nodes, the same is true for the node degrees in the augmented factor graph. As a result, the inference time per variable node remains constant, because the inference only requires information from the node's $K$-hop neighbourhood, the number of which also does not increase with the total number of nodes in the graph. This implies that the computational complexity of inferring all of the state variables is $\mathcal{O}(n)$. Due to the over-smoothing problem in GNNs \cite{Chen_2020_oversmoothing}, a small value is assigned to $K$, thus not affecting the computational complexity of the inference.

The inference should be computationally and geographically distributed, to make the best use of the proposed approach for the large scale power systems in which the communication delays between the PMUs and the central processing unit are the obstacle for full utilisation of the PMU large sampling rates. The distributed implementation is possible as long as all of the measurements within the $K$-hop neighbourhood in the augmented factor graph are fed into the computational module that generates the predictions. For arbitrary $K$, PMUs required for the inference will be physically located within the $\lceil K/2 \rceil$-hop neighbourhood of the power system bus.

\section{Numerical Results}
In this section, we describe the training setups used for augmented factor graph-based GNN models and assess the prediction quality of the trained models in various test scenarios\footnote{Source code is available online at https://github.com/ognjenkundacina.}. Training, validation, and test sets are obtained by labeling various samples of measurement means, variances, and covariances with the WLS solutions of the system described in (\ref{SE_system_of_lin_eq}). The measurements are obtained by adding Gaussian noise to the exact power flow solutions, with each power flow calculation executed with a different load profile. GNN models are tested in situations in which the number of PMUs is optimal (with minimal measurement redundancy under which the WLS SE provides a solution), in underdetermined scenarios, as well as in scenarios with maximal measurement redundancy. We used the IGNNITION framework \cite{pujolperich2021ignnition} for building and utilising GNN models, with the hyperparameters presented in Table~\ref{tbl_hyperparameters} used in all of the experiments. 

\begin{table}[!t]
\caption{List of GNN hyperparameters.}
\label{tbl_hyperparameters}
\begin{center}
    \begin{tabular}{ | l | c | }
        \hline
        \textbf{Hyperparameters} & \textbf{Values} \\ 
        \hline
        Node embedding size $s$ & $64$ \\
        \hline
        Learning rate & $4\times 10^{-4}$ \\
        \hline
        Minibatch size $B$ & $32$ \\
        \hline
        Number of GNN layers $K$ & $4$ \\
        \hline
        Activation functions & ReLU \\
        \hline
        Optimizer & Adam \\
        \hline
        Batch normalization type & Mean \\
        \hline
    \end{tabular}
\end{center}
\vspace{-6mm}
\end{table}

\subsection{Power system with optimally placed PMUs}
\label{subsec:optPlacedPMUs}

In this subsection, we will run a series of experiments on the IEEE 30-bus power system, each using different measurement variances for training and test set generation, with values $10^{-5}$, $10^{-3}$, $10^{-2}$, $10^{-1}$, and $0.5$. The training set for each of the experiments consists of $10000$ samples, while the validation and test set both contain $100$ samples. The number and the positions on PMUs are fixed and determined using the optimal PMU placement algorithm \cite{optimalPMUPlacement}, which finds the smallest set of PMU measurements that make the system observable. The algorithm has resulted in a total of 10 PMUs and $50$ measurement phasors, $10$ of which are the voltage phasors and the rest are the current phasors.

Table \ref{tbl_MSEsIEEE30} displays test set results in the form of average mean square errors (MSEs) between GNN predictions and test set labels for all experiments, distinguished by the measurement variances column. These results are compared with the average MSE between the labels and the approximative WLS SE solutions defined in Sec. \ref{sec:lse}. The results show that for systems with optimally placed PMUs and low measurement variances, GNN predictions have very small deviations from the exact WLS SE, despite being outperformed by the approximative WLS SE. For higher measurement variances, GNN has a lower estimation error than the approximative WLS SE while also having lower computational complexity in all of the cases.

\begin{table}[htbp]
\caption{Comparison of GNN and approximative SE test set MSEs for various measurement variances .}
\begin{center}
    \begin{tabular}{ | c | c | c | }
        \hline
        \textbf{Meas. variances} & \textbf{GNN}  & \textbf{Approx. SE} \\
        \hline
        $10^{-5}$ & $2.48\times 10^{-6}$ & $1.87\times 10^{-8}$ \\
        \hline
        $10^{-3}$ & $8.21\times 10^{-6}$ & $2.25\times 10^{-3}$ \\
        \hline
        $10^{-2}$ & $2.25\times 10^{-5}$ & $3.66\times 10^{-5}$ \\
        \hline
        $10^{-1}$ & $7.47\times 10^{-4}$ & $2.27\times 10^{-3}$ \\
        \hline
        $0.5$ & $6.71\times 10^{-3}$ & $3.66\times 10^{-2}$ \\
        \hline
    \end{tabular}
\label{tbl_MSEsIEEE30}
\end{center}
\end{table}

\begin{figure}[htbp]
    \centering
    \pgfplotstableread[col sep = comma,]{data/PredictionsPerNode0Excluded.csv}\CSVPredVSObservedZeroExcludedEdgesVarNodes
    
        \begin{tikzpicture}
        
        \begin{groupplot}[group style={group size=1 by 2, vertical sep=0.5cm}, ]
        \nextgroupplot[
        yscale=0.55,
        legend cell align={left},
        legend columns=1,
        legend style={
          fill opacity=0.8,
          font=\small,
          draw opacity=1,
          text opacity=1,
          at={(0.45,0.05)},
          anchor=south,
          draw=white!80!black
        },
        scaled x ticks=manual:{}{\pgfmathparse{#1}},
        tick align=outside,
        tick pos=left,
        x grid style={white!69.0196078431373!black},
        xmin=-1.45, xmax=30.45,
        xtick style={color=black},
        xticklabels={},
        y grid style={white!69.0196078431373!black},
        ylabel={\(\displaystyle \Re({V}_i)\)},
        ymin=0.93, ymax=1.08,
        ytick style={color=black},
        xmajorgrids=true,
        ymajorgrids=true,
        grid style=dashed
        ]
        \addplot [red, mark=square*, mark options={scale=0.5}, mark repeat=2,mark phase=1]
        table [x expr=\coordindex, y={predictionsRE}]{\CSVPredVSObservedZeroExcludedEdgesVarNodes};
        \addlegendentry{Predictions}
        
        \addplot [blue, mark=square*, mark options={scale=0.5}, mark repeat=2,mark phase=2]
        table [x expr=\coordindex, y={true_valuesRE}]{\CSVPredVSObservedZeroExcludedEdgesVarNodes};
        \addlegendentry{Labels}
        
        
        
        
        \nextgroupplot[
        yscale=0.55,
        tick align=outside,
        tick pos=left,
        x grid style={white!69.0196078431373!black},
        xlabel={Bus index ($i$)},
        xmin=-1.45, xmax=30.45,
        xtick style={color=black},
        y grid style={white!69.0196078431373!black},
        ylabel={\(\displaystyle \Im({V}_i)\)},
        ymin=-0.35, ymax=0.05,
        ytick style={color=black},
        xmajorgrids=true,
        ymajorgrids=true,
        grid style=dashed
        ]
        
        \addplot [red, mark=square*, mark options={scale=0.5}, mark repeat=2,mark phase=1]
        table [x expr=\coordindex, y={predictionsIM}]{\CSVPredVSObservedZeroExcludedEdgesVarNodes};
        
        \addplot [blue, mark=square*, mark options={scale=0.5}, mark repeat=2,mark phase=2]
        table [x expr=\coordindex, y={true_valuesIM}]{\CSVPredVSObservedZeroExcludedEdgesVarNodes};
        
                
        \addplot [brown]
        table [x expr=\coordindex, y={approxSeIM}]{\CSVPredVSObservedZeroExcludedEdgesVarNodes};
        
        
        \end{groupplot}
        
        \end{tikzpicture}
    \caption{GNN predictions, approximative WLS SE solutions, and labels for one test example with no measurement phasors removed.}
    \label{PredictionsPerNode0Excluded}
\end{figure}
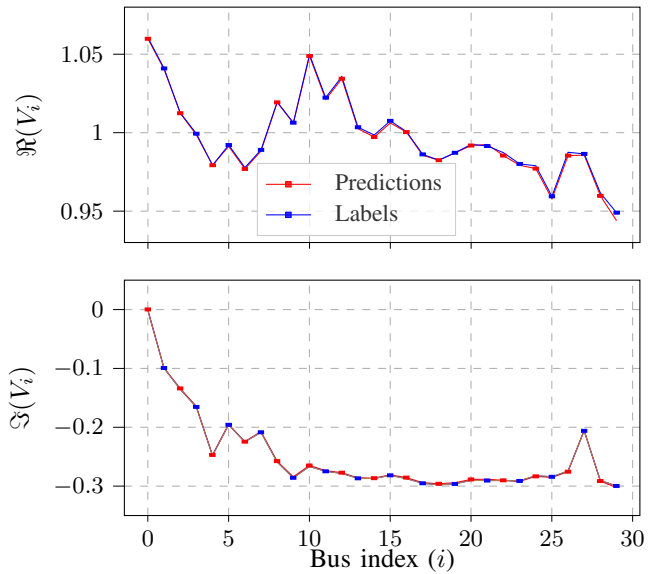

In Fig.~\ref{PredictionsPerNode0Excluded} we present the predictions and the labels for each of the variable nodes for one of the test samples. The results for the real parts of complex node voltages in the upper plot and the imaginary parts in the lower plot indicate that GNNs can be used as accurate SE approximators. 

\begin{figure}[htbp]
    \centering
    \pgfplotstableread[col sep = comma,]{data/PredictionsPerNode2PMUExcluded.csv}\CSVPredVSObservdTwoPMUExcludedEdgesVarNodes
    
        \begin{tikzpicture}
        
        \begin{groupplot}[group style={group size=1 by 2, vertical sep=0.5cm}, ]
        \nextgroupplot[
        yscale=0.55,
        legend cell align={left},
        legend columns=1,
        legend style={
          fill opacity=0.8,
          font=\small,
          draw opacity=1,
          text opacity=1,
          at={(0.195,17.149)},
          anchor=south,
          draw=white!80!black,
          nodes={scale=0.98, transform shape}
        },
        scaled x ticks=manual:{}{\pgfmathparse{#1}},
        tick align=outside,
        tick pos=left,
        x grid style={white!69.0196078431373!black},
        xmin=-1.45, xmax=30.45,
        xtick style={color=black},
        xticklabels={},
        y grid style={white!69.0196078431373!black},
        ylabel={\(\displaystyle \Re({V}_i)\)},
        ymin=0.93, ymax=1.07,
        ytick style={color=black}
        ]
        \addplot [red, mark=square*, mark options={scale=0.5}, mark repeat=2,mark phase=1]
        table [x expr=\coordindex, y={predictionsRE}]{\CSVPredVSObservdTwoPMUExcludedEdgesVarNodes};
        \addlegendentry{Predictions}
        
        \addplot [blue, mark=square*, mark options={scale=0.5}, mark repeat=2,mark phase=2]
        table [x expr=\coordindex, y={true_valuesRE}]{\CSVPredVSObservdTwoPMUExcludedEdgesVarNodes};
        \addlegendentry{Labels}
        
        
        
        \addplot [very thin, black, dashed, forget plot]
        table {%
        14 -2.0
        14 2.0
        };
        \addplot [very thin, black, dashed, forget plot]
        table {%
        11 -2.0
        11 2.0
        };
        \addplot [very thin, black, dashed, forget plot]
        table {%
        13 -2.0
        13 2.0
        };
        \addplot [very thin, black, dashed, forget plot]
        table {%
        22 -2.0
        22 2.0
        };
        \addplot [very thin, black, dashed, forget plot]
        table {%
        17 -2.0
        17 2.0
        };
        
        \addplot [very thin, black, dashed, forget plot]
        table {%
        18 -2.0
        18 2.0
        };

        \nextgroupplot[
        yscale=0.55,
        tick align=outside,
        tick pos=left,
        x grid style={white!69.0196078431373!black},
        xlabel={Bus index ($i$)},
        xmin=-1.45, xmax=30.45,
        xtick style={color=black},
        y grid style={white!69.0196078431373!black},
        ylabel={\(\displaystyle \Im({V}_i)\)},
        ymin=-0.35, ymax=0.08,
        ytick style={color=black}
        ]
        
        \addplot [red, mark=square*, mark options={scale=0.5}, mark repeat=2,mark phase=1]
        table [x expr=\coordindex, y={predictionsIM}]{\CSVPredVSObservdTwoPMUExcludedEdgesVarNodes};
        
        \addplot [blue, mark=square*, mark options={scale=0.5}, mark repeat=2,mark phase=2]
        table [x expr=\coordindex, y={true_valuesIM}]{\CSVPredVSObservdTwoPMUExcludedEdgesVarNodes};
        
        
        
        \addplot [very thin, black, dashed, forget plot]
        table {%
        14 -2
        14 2
        };
        \addplot [very thin, black, dashed, forget plot]
        table {%
        11 -2.0
        11 2.0
        };
        \addplot [very thin, black, dashed, forget plot]
        table {%
        13 -2.0
        13 2.0
        };
        \addplot [very thin, black, dashed, forget plot]
        table {%
        22 -2.0
        22 2.0
        };
        \addplot [very thin, black, dashed, forget plot]
        table {%
        17 -2.0
        17 2.0
        };
        
        \addplot [very thin, black, dashed, forget plot]
        table {%
        18 -2.0
        18 2.0
        };
        
        \end{groupplot}
        
        \end{tikzpicture}

    \caption{GNN predictions and labels for one test example with phasors from two neighboring PMUs removed.}
    \label{PredictionsPerNode2PMUExcluded}
\end{figure}
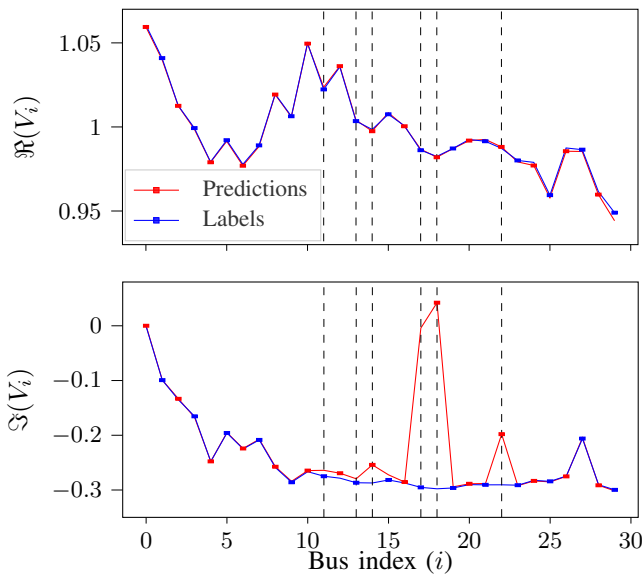

To further analyse the robustness of the proposed model, we test it by excluding several measurement phasors from the previously used test samples, making the system of equations that describes the SE problem underdetermined. Excluding the measurement phasor from the test sample is realised by removing its real and imaginary parts from the input data, which is equivalent to removing two factor nodes from the augmented factor graph. Using the previously used 100-sample test set we create a new test set, by removing selected measurement phasors from each sample, while preserving the same labels obtained as SE solutions of the system with all of the measurements present. As an example, we observe the predictions for the scenario where two neighboring PMUs fail to deliver measurements to the state estimator, hence all of the 8 measurement phasors associated with the removed PMUs are excluded from the GNN inputs. Average MSE for the test set of 100 samples created by removing these measurements from the original test set used in this section equals $3.45\times 10^{-3}$. Predictions and labels per variable node index for one test set sample are shown in Fig \ref{PredictionsPerNode2PMUExcluded}, in which vertical dashed lines indicate indices of the variable nodes within the $2$-hop neighbourhood of the removed PMUs. We can observe that significant deviations from the labels occur for some of the neighboring buses, while GNN predictions are a decent fit for the remaining node labels. The proposed model demonstrated the ability to sustain the error in the neighborhood of the malfunctioning PMU, as well as robustness in scenarios not solvable using the standard WLS approaches.

\subsection{Scalability and sample efficiency analysis}

To assess how does the our approach scale to larger power systems, we compare the GNN model results with the approximative WLS SE for IEEE 118-bus, IEEE 300-bus and the ACTIVSg 2000-bus power systems \cite{ACTIVSg2000}. In contrast to the example in Sec. \ref{subsec:optPlacedPMUs}, we used maximal measurement redundancies, ranging from $3.73$ to $4.21$. We chose an unrealistically high measurement variance with a value of $0.5$ as an edge case, as we have seen in the previous subsection that GNN predictions demonstrated high accuracy for low measurement variances. To demonstrate how sample efficient is the proposed method, we create training sets with sizes of $10$, $100$, $1000$ and $10000$ samples for each of the mentioned power systems, alongside with previously used IEEE 30-bus power system. We trained an instance of a GNN model for the generated training sets\footnote{\label{trainingFeasibility}Training the model for the ACTIVSg 2000-bus power systems on $10000$ samples could not be performed due to requirements that exceeded our hardware configuration (Intel Core(TM) i7-11800H 2.30 GHz CPU, and 16 GB of RAM). Optimising the batch loading of the training examples would make the training process feasible for large datasets even on a modest hardware setup.}, and each training converged within 200 epochs. In Table \ref{tbl_all_schemes}, we report the MSEs between the labels and the outputs on 100-sample sized test sets for all of the trained models and the approximative WLS SE. Results indicate that even the GNN models trained on a small dataset outperform the approximative WLS SE, with the exception of the model trained for IEEE 30-bus power system on $10$ samples. It can be concluded that the quality of GNN model's predictions improves with the increase in the amount of training data, and the emboldened results for the best models have significantly smaller MSE compared to approximative WLS SE. Although we used randomly generated training sets, narrowing the learning space by carefully selecting training samples based on historical load consumption data could result in even better results on small datasets. 

We emphasise that the number of trainable parameters remains nearly constant when the number of the power system buses increases. The exception is the number of input neurons for variable node binary index encoding, which grows logarithmically with the number of variable nodes but is insignificant compared to the total number of GNN parameters.

\begin{table*}[htbp]
\caption{Test set results for various power systems and training set sizes.}
\begin{center}
    \begin{tabular}{ | c | c | c | c | c | c | }
        \hline
        \textbf{Power system} & \textbf{10 samples} & \textbf{100 samples} & \textbf{1000 samples} & \textbf{10000 samples} & \textbf{Approx. SE}\\
        \hline
        IEEE 30 & $3.77\times 10^{-2}$ & $3.44\times 10^{-2}$ & $1.98\times 10^{-2}$ & \boldmath{$6.71\times 10^{-3}$} & $3.66\times 10^{-2}$ \\
        \hline
        IEEE 118 & $9.89\times 10^{-3}$ & $9.01\times 10^{-3}$ & $9.00\times 10^{-3}$ & \boldmath{$8.81\times 10^{-3}$} & $4.15\times 10^{-2}$ \\
        \hline
        IEEE 300 & $2.34\times 10^{-2}$ & $1.61\times 10^{-2}$ & $1.11\times 10^{-2}$ & \boldmath{$1.07\times 10^{-2}$} & $5.14\times 10^{-2}$ \\
        \hline
        ACTIVSg 2000 & $1.25\times 10^{-2}$ & $5.011\times 10^{-3}$ & \boldmath{$4.68\times 10^{-3}$} & 
        No results & $3.79\times 10^{-2}$ \\
        \hline

    \end{tabular}
\label{tbl_all_schemes}
\end{center}
\end{table*}

\subsection{Robustness to outliers}
We evaluate the robustness of the proposed model to outliers among input data using the same experiment setup described in Sec. \ref{subsec:optPlacedPMUs} and test samples generated with measurement variance $10^{-5}$. In every test set sample, we randomly replace an existing measurement with a value generated with a variance $1600$ times higher than the original one and transformed it into rectangular coordinates. For labels, we use WLS SE solutions with no outliers among the inputs. Table \ref{tbl_outliers} displays MSEs between the labels and the outputs of three different approaches, along with the results of the WLS SE and the approximative WLS SE algorithm for the same test set.

The first approach, which uses the already trained model from Sec. \ref{subsec:optPlacedPMUs}, results in high MSE of the prediction on the test set with outliers. The main contribution to the MSE are large mismatches from the labels in the $K$-hop neighbourhood of the outlier, caused by the fact that the ReLU activation function does not limit its inputs during neighbourhood aggregation. To cope with this problem, as the second approach, we trained a GNN model with the same architecture as the previous one, but which uses the tanh (hyperbolic tangent) activation function instead of ReLU. As presented in Table \ref{tbl_outliers}, this approach resulted in a test set MSE that is significantly lower than the proposed GNN with ReLU activations, WLS SE, and the approximative WLS SE as well. The saturation effect of tanh prevents high values stemming from outliers from propagating through the neural network but also diminishes the training quality due to the vanishing gradient problem. Namely, all of the experiments we conducted under the same conditions for GNN with tanh activations required more epochs to converge to the solution with lower prediction qualities compared to the GNN with ReLU activations. As a third approach, we propose training a GNN model with ReLU activations on a dataset in which half of the samples contain outliers, which are generated in the same manner as the test samples used in this subsection. This approach is the most effective because the GNN learns to neutralise the effect of unexpected inputs from the dataset examples while maintaining accurate predictions in the absence of outliers in the input data. We note that these are only preliminary efforts to make the GNN model robust to outliers, and that future research could combine ideas from standard bad data processing methods in SE with the proposed GNN approach.

\begin{table}[htbp]
\caption{A comparison of the results of various approaches for test sets containing outliers.}
\begin{center}
    \begin{tabular}{ | c | c | }
        \hline
        \textbf{Approach} & \textbf{Test set MSE} \\
        \hline
        GNN & $1.60\times 10^{3}$ \\
        \hline
        GNN with tanh & $2.39\times 10^{-2}$ \\
        \hline
        GNN trained with outliers & $7.99\times 10^{-6}$ \\
        \hline
        WLS SE & $1.41\times 10^{-1}$ \\
        \hline
        Approximative WLS SE & $1.35\times 10^{-1}$ \\
        \hline
    \end{tabular}
\label{tbl_outliers}
\end{center}
\end{table}

\section{Conclusions}
In this research, we investigate how GNNs can be used as fast solvers of linear SE with PMUs. The proposed graph attention network based model, specialised for newly introduced heterogeneous augmented factor graphs, recursively propagates the inputs from the factor nodes to generate the predictions in the variable nodes based on the WLS SE solutions. Evaluating the trained model on the unseen data samples confirms that the proposed GNN approach can be used as a very accurate approximator of the linear WLS SE solutions, with the added benefit of linear computational complexity at the inference time. The model is robust in unobservable scenarios that are not solvable using standard WLS SE methods, such as when individual phasor measurements or entire PMUs fail to deliver measurement data to the proposed SE solver. Furthermore, when measurement variances are high or outliers are present in the input data, the GNN model outperforms the approximate WLS SE. The proposed approach demonstrates scalability as well as sample efficiency when tested on power systems of various sizes, as it predicts the state variables well even when trained on a small number of randomly generated samples. 

While we focused on solving linear SE with PMUs, the proposed learning framework, graph augmentation techniques, and conclusions could be applied to a variety of SE formulations. One of them could be a GNN-based SE for highly unobservable distribution systems, motivated by the GNN's ability to provide relevant solutions in underdetermined scenarios.


\section{Acknowledgment}
This paper has received funding from the European Union's Horizon 2020 research and innovation programme under Grant Agreement number 856967.

\bibliographystyle{IEEEtran}
\bibliography{cite}

\end{document}

%% file: acronyms.tex
\newacronym{se}{SE}{State Estimation}
\newacronym{bp}{BP}{Belief Propagation}

\usepackage{xspace}

%% file: journal.bbl
\begin{thebibliography}{10}
\providecommand{\url}[1]{#1}
\csname url@samestyle\endcsname
\providecommand{\newblock}{\relax}
\providecommand{\bibinfo}[2]{#2}
\providecommand{\BIBentrySTDinterwordspacing}{\spaceskip=0pt\relax}
\providecommand{\BIBentryALTinterwordstretchfactor}{4}
\providecommand{\BIBentryALTinterwordspacing}{\spaceskip=\fontdimen2\font plus
\BIBentryALTinterwordstretchfactor\fontdimen3\font minus
  \fontdimen4\font\relax}
\providecommand{\BIBforeignlanguage}[2]{{%
\expandafter\ifx\csname l@#1\endcsname\relax
\typeout{** WARNING: IEEEtran.bst: No hyphenation pattern has been}%
\typeout{** loaded for the language `#1'. Using the pattern for}%
\typeout{** the default language instead.}%
\else
\language=\csname l@#1\endcsname
\fi
#2}}
\providecommand{\BIBdecl}{\relax}
\BIBdecl

\bibitem{monticelli2000SE}
A.~Monticelli, ``Electric power system state estimation,'' \emph{Proceedings of
  the IEEE}, vol.~88, no.~2, pp. 262--282, 2000.

\bibitem{exposito}
A.~Gomez-Exposito, A.~Abur, P.~Rousseaux, A.~de~la Villa~Jaen, and
  C.~Gomez-Quiles, ``On the use of {PMU}s in power system state estimation,''
  \emph{Proceedings of the 17th PSCC}, 2011.

\bibitem{pmlr-v70-gilmer17a}
J.~Gilmer, S.~S. Schoenholz, P.~F. Riley, O.~Vinyals, and G.~E. Dahl, ``Neural
  message passing for quantum chemistry,'' in \emph{Proceedings of the 34th
  International Conference on Machine Learning}.\hskip 1em plus 0.5em minus
  0.4em\relax PMLR, 06--11 Aug 2017, pp. 1263--1272.

\bibitem{GraphRepresentationLearningBook}
W.~L. Hamilton, ``Graph representation learning,'' \emph{Synthesis Lectures on
  Artificial Intelligence and Machine Learning}, vol.~14, no.~3, pp. 1--159,
  2020.

\bibitem{velickovic2018graph}
P.~Veli{\v{c}}kovi{\'{c}}, G.~Cucurull, A.~Casanova, A.~Romero, P.~Li{\`{o}},
  and Y.~Bengio, ``{Graph Attention Networks},'' \emph{International Conference
  on Learning Representations}, 2018.

\bibitem{antona2018PSUncertainty}
G.~D’Antona, ``Power system static-state estimation with uncertain network
  parameters as input data,'' \emph{IEEE Trans. Instrum. Meas.}, vol.~65,
  no.~11, pp. 2485--2494, 2016.

\bibitem{gong2019EdgeFeatures}
L.~Gong and Q.~Cheng, ``Exploiting edge features for graph neural networks,''
  in \emph{Proc. IEEE/CVF CVPR}, 2019, pp. 9203--9211.

\bibitem{zhang2019}
L.~Zhang, G.~Wang, and G.~B. Giannakis, ``Real-time power system state
  estimation and forecasting via deep unrolled neural networks,'' \emph{IEEE
  Trans. Signal Process.}, vol.~67, no.~15, pp. 4069--4077, 2019.

\bibitem{zamzam2019}
A.~S. Zamzam, X.~Fu, and N.~D. Sidiropoulos, ``Data-driven learning-based
  optimization for distribution system state estimation,'' \emph{IEEE Trans.
  Power Syst.}, vol.~34, no.~6, pp. 4796--4805, 2019.

\bibitem{donon2019graphneuralsolver}
B.~Donon, B.~Donnot, I.~Guyon, and A.~Marot, ``Graph neural solver for power
  systems,'' in \emph{2019 International Joint Conference on Neural Networks
  (IJCNN)}, 2019, pp. 1--8.

\bibitem{bolz2019PFapproximator}
V.~Bolz, J.~Rueß, and A.~Zell, ``Power flow approximation based on graph
  convolutional networks,'' in \emph{Proc. IEEE ICMLA}, 2019, pp. 1679--1686.

\bibitem{wang2020ProbabPF}
D.~Wang, K.~Zheng, Q.~Chen, G.~Luo, and X.~Zhang, ``Probabilistic power flow
  solution with graph convolutional network,'' in \emph{2020 IEEE PES
  Innovative Smart Grid Technologies Europe (ISGT-Europe)}, 2020, pp. 650--654.

\bibitem{donon2020graphneuralsolverJOURNAL}
B.~Donon, R.~Clément, B.~Donnot, A.~Marot, I.~Guyon, and M.~Schoenauer,
  ``Neural networks for power flow: Graph neural solver,'' \emph{Electric Power
  Systems Research}, vol. 189, p. 106547, 2020.

\bibitem{pagnier2021physicsinformed}
L.~Pagnier and M.~Chertkov, ``Physics-informed graphical neural network for
  parameter \& state estimations in power systems,'' 2021.

\bibitem{Yang2022RobustPSSEDataDrivenPriors}
Q.~Yang, A.~Sadeghi, and G.~Wang, ``Data-driven priors for robust psse via
  gauss-newton unrolled neural networks,'' \emph{IEEE Journal on Emerging and
  Selected Topics in Circuits and Systems}, vol.~PP, pp. 1--1, 01 2022.

\bibitem{TGCN_SE_2021}
M.~J. Hossain and M.~Rahnamay–Naeini, ``State estimation in smart grids using
  temporal graph convolution networks,'' in \emph{2021 North American Power
  Symposium (NAPS)}, 2021, pp. 01--05.

\bibitem{Satorras2021NeuralEB}
V.~G. Satorras and M.~Welling, ``Neural enhanced belief propagation on factor
  graphs,'' in \emph{Proc. AISTATS}, 2021.

\bibitem{Kschischang2001FactorGraphs}
F.~Kschischang, B.~Frey, and H.-A. Loeliger, ``Factor graphs and the
  sum-product algorithm,'' \emph{IEEE Transactions on Information Theory},
  vol.~47, no.~2, pp. 498--519, 2001.

\bibitem{kundacina2022state}
O.~Kundacina, M.~Cosovic, and D.~Vukobratovic, ``State estimation in electric
  power systems leveraging graph neural networks,'' in \emph{International
  Conference on Probabilistic Methods Applied to Power Systems (PMAPS)}, 2022.

\bibitem{phadke}
J.~De~La~Ree, V.~Centeno, J.~S. Thorp, and A.~G. Phadke, ``Synchronized phasor
  measurement applications in power systems,'' \emph{IEEE Trans. Smart Grid},
  vol.~1, no.~1, pp. 20--27, 2010.

\bibitem{zhou2016phasorsMeasSE}
M.~Zhou, V.~Centeno, J.~Thorp, and A.~Phadke, ``An alternative for including
  phasor measurements in state estimators,'' \emph{IEEE Trans. Power Syst.},
  vol.~21, no.~4, pp. 1930--1937, 2006.

\bibitem{KipfW17_GCN}
T.~N. Kipf and M.~Welling, ``Semi-supervised classification with graph
  convolutional networks,'' in \emph{5th International Conference on Learning
  Representations, {ICLR}}, 2017.

\bibitem{cosovic2019bpse}
M.~Cosovic and D.~Vukobratovic, ``Distributed {G}auss–{N}ewton method for
  state estimation using belief propagation,'' \emph{IEEE Trans. Power Syst.},
  vol.~34, no.~1, pp. 648--658, 2019.

\bibitem{Chen_2020_oversmoothing}
D.~Chen, Y.~Lin, W.~Li, P.~Li, J.~Zhou, and X.~Sun, ``Measuring and relieving
  the over-smoothing problem for graph neural networks from the topological
  view,'' \emph{Proceedings of the AAAI Conference on Artificial Intelligence},
  vol.~34, no.~04, pp. 3438--3445, Apr. 2020.

\bibitem{pujolperich2021ignnition}
D.~Pujol-Perich, J.~Suárez-Varela, M.~Ferriol, S.~Xiao, B.~Wu,
  A.~Cabellos-Aparicio, and P.~Barlet-Ros, ``Ignnition: Bridging the gap
  between graph neural networks and networking systems,'' 2021.

\bibitem{optimalPMUPlacement}
B.~Gou, ``Optimal placement of pmus by integer linear programming,'' \emph{IEEE
  Trans. Power Syst.}, vol.~23, pp. 1525 -- 1526, 09 2008.

\bibitem{ACTIVSg2000}
A.~B. Birchfield, T.~Xu, K.~M. Gegner, K.~S. Shetye, and T.~J. Overbye, ``Grid
  structural characteristics as validation criteria for synthetic networks,''
  \emph{IEEE Trans. Power Syst.}, vol.~32, no.~4, pp. 3258--3265, 2017.

\end{thebibliography}
